\documentclass[10pt,twocolumn,letterpaper]{article}

\usepackage{iccv}
\usepackage{times}
\usepackage{epsfig}
\usepackage{graphicx}
\usepackage{amsmath}
\usepackage{amssymb}
\usepackage{multirow}

\usepackage[pagebackref=true,breaklinks=true,letterpaper=true,colorlinks,bookmarks=false]{hyperref}

\iccvfinalcopy % *** Uncomment this line for the final submission

\def\iccvPaperID{2601} % *** Enter the ICCV Paper ID here

\ificcvfinal\fi
\begin{document}

%%%%%%%%% TITLE
\title{Temporal Context Network for Activity Localization in Videos}

\author{Xiyang Dai$^1 \quad\quad$ Bharat Singh$^1 \quad\quad$ Guyue Zhang$^2$ \\
$^1$University of Maryland\\
College Park, MD\\
{\tt\small {xdai, bharat, lsd}@umiacs.umd.edu}
\and
Larry S. Davis$^1 \qquad$  Yan Qiu Chen$^2$\\
$^2$Fudan University\\
Shanghai, China\\
{\tt\small {guyuezhang13, chenyq}@fudan.edu.cn}
}

\maketitle
\thispagestyle{empty}

%%%%%%%%% ABSTRACT
\begin{abstract}
   We present a Temporal Context Network (TCN) for precise temporal localization of human activities. Similar to the Faster-RCNN architecture, proposals are placed at equal intervals in a video which span multiple temporal scales. We propose a novel representation for ranking these proposals. Since pooling features only inside a segment is not sufficient to predict activity boundaries, we construct a representation which explicitly captures context around a proposal for ranking it. For each temporal segment inside a proposal, features are uniformly sampled at a pair of scales and are input to a temporal convolutional neural network for classification. After ranking proposals, non-maximum suppression is applied and classification is performed to obtain final detections. TCN outperforms state-of-the-art methods on the ActivityNet dataset and the THUMOS14 dataset.
\end{abstract} 

%%%%%%%%% BODY TEXT
\section{Introduction}
%problem and motivation
Recognizing actions and activities in videos is a long studied problem in computer vision \cite{bobick2001recognition,haritaoglu2000w,bregler1997learning}. An action is defined as a short duration movement such as jumping, throwing, kicking. In contrast, activities are more complex. An activity has a beginning, which is triggered by an action or an event, which involves multiple actions, and an end, which involves another action or an event. For example, an activity like ``assembling a furniture" could start with unpacking boxes, continue by putting different parts together and end when the furniture is ready. Since videos can be arbitrarily long, they may contain multiple activities and therefore, temporal localization is needed. Detecting human activities in videos has several applications in content based video retrieval for web search engines, reducing the effort required to browse through lengthy videos, monitoring suspicious activity in video surveillance etc. While localizing objects in images is an extensively studied problem, localizing activities has received less attention. This is primarily because performing localization in videos is computationally expensive \cite{escorcia2016daps} and well annotated large datasets \cite{caba2015activitynet} were unavailable until recently. 

Current object detection pipelines have three major components - proposal generation, object classification and bounding box refinement \cite{ren2015faster}. In \cite{escorcia2016daps, shou2016action} this pipeline was adopted for deep learning based action detection as well. LSTM is used to embed a long video into a single feature vector which is then used to score different segment proposals in the video \cite{escorcia2016daps}. While a LSTM is effective for capturing local context in a video \cite{Singh_2016_CVPR}, learning to predict the start and end positions for all activity segments using the hidden state of a LSTM is challenging. In fact, in our experiments we show that even a pre-defined set of proposals at multiple scales obtains better recall than the temporal segments predicted by a LSTM on the ActivityNet dataset. 
% \begin{figure}[tb]
% \center
% \includegraphics[width=0.95\linewidth]{figures/intro.png}
% \caption{ A video sequence depicted as a sequence of images, where the blue bar represents the ground truth while the green bar represents proposals. Notice that (b) is a bad proposal while (a) is not.
% }
% \label{fig:intro}
% \end{figure}

\begin{figure*}
    \center
    \includegraphics[width=0.8\linewidth]{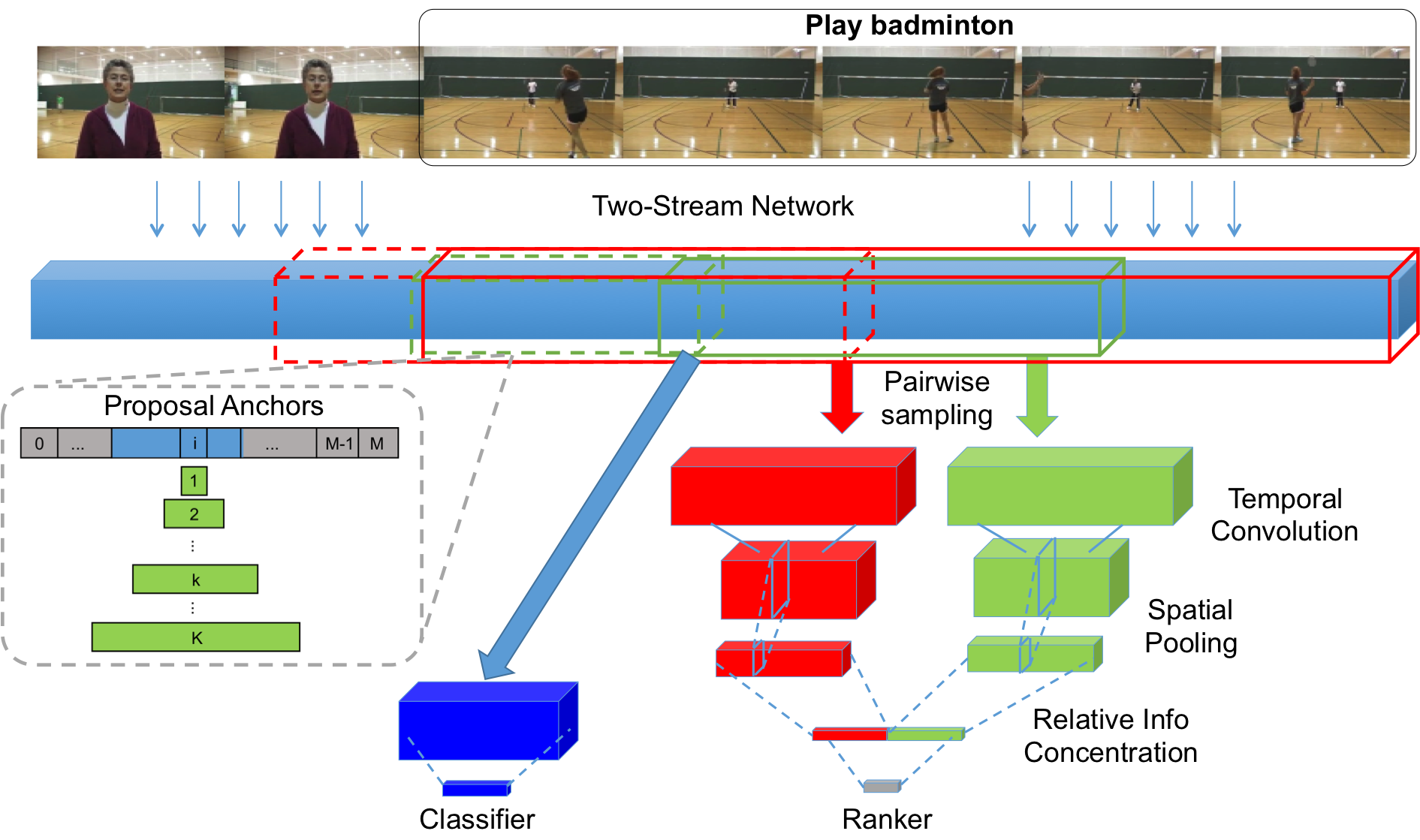}
    \caption{Given a video, a two stream network is used to extract features. A pair-wise sampling layer samples features at two different resolutions to construct the feature representation for a proposal. This pairwise sampling helps to obtain a better proposal ranking. A typical sliding window approach (Green line box) can miss the context boundary information when it lies inside the activity. However, the proposed pairwise sampling with a larger context window (Red line box) will capture such information and yield better proposal ranking. These pair-wise features are then input to a ranker which selects proposals for classification. The green boxes on the left represent K different proposals which are placed uniformly in a video.}
    \label{fig:demo}
\end{figure*}

In \cite{shou2016action}, a ranker was learned on multiple segments of a video based on overlap with ground truth segments. However, a feature representation which does not integrate information from a larger temporal scale than a proposal lacks sufficient information to predict whether a proposal is a good candidate or not. For example, in Figure \ref{fig:demo}, the red and green solid segments are two proposals which are both completely included within an activity. While the red segment is a good candidate, the green is not. So, although a single scale representation for a segment captures sufficient information for recognition, it is inadequate for detection. To capture information for predicting activity boundaries, we propose to explicitly sample features both at the scale of the proposal and also at a higher scale while ranking proposals. We experimentally demonstrate that this has significant impact on performance when ranking temporal activity proposals.

By placing proposals at equal intervals in a video which span multiple temporal scales, we construct a set of proposals which are then ranked using features sampled from a pair of scales. A temporal convolution network is applied over these features to learn background and foreground probabilities. The top ranked proposals are then input to a classification network which assigns individual class probabilities to each segment proposal.
%how is it different

%summarize novelty

\section{Related Work} 
Wang and Schmidt \cite{wang2011action} introduced Dense Trajectories (DT), which have been widely applied in various video recognition algorithms. For trimmed activity recognition, extracting dense trajectories and encoding them by using Fisher Vectors has been widely used \cite{atmosukarto2012trajectory, wang2013action, jiang2014thumos, heilbron2014camera, peng2016bag, wang2016improving}. For action detection, \cite{yuan2016temporal} constructed a pyramid of score distribution features (PSDF) as a representation for ranking segments of a video  in a dense trajectories based pipeline. However, for large datasets, these methods require significant computational resources to extract features and build the feature representation after features are extracted. Because deep learning based methods provide better accuracy with much less computation, hand-crafted features have become less popular.

For object detection in images, proposals are a critical elements for obtaining efficient and accurate detections \cite{russakovsky2015imagenet, ren2015faster}. Motivated by this approach, Jain et al. \cite{jain2014action} introduced action proposals which extends object proposals to videos. For spatio-temporal localization of actions, multiple methods use spatio-temporal region proposals \cite{gkioxari2015finding,oneata2014spatio,gemert2015apt,yu2015fast}. However, these methods are typically applied to datasets containing short videos, and hence the major focus has been on spatial localization rather than temporal localization. Moreover, spatio-temporal localization requires training data containing frame level bounding box annotations. For many applications, simply labeling the action boundaries in the video is sufficient, which is a significantly less cumbersome annotation task.

Very recently, studies focusing on temporal segments which contain human actions have been introduced \cite{mettes2015bag, caba2016fast, shou2016action, ma2016learning, Singh_2016_CVPR}. Similar to grouping techniques for retrieving object proposals, Heilbron et al. \cite{caba2016fast} used a sparse dictionary to encode discriminative information for a set of action classes. Mettes et al. \cite{mettes2015bag} introduced a fragment hierarchy based on semantic visual similarity of contiguous frames by hierarchical clustering, which was later used to efficiently encode temporal segments in unseen videos. In \cite{Singh_2016_CVPR}, a multi-stream RNN was employed along with tracking to generate frame level predictions to which simple grouping was applied at multiple detection thresholds for obtaining detections.

Methods using category-independent classifiers to obtain many segments in a long video are more closely related to our approach. For example, Shou et al. \cite{shou2016action} exploit three segment-based 3D ConvNets: a proposal network for identifying candidate clips that may contain actions, a classification network for learning a classification model and a localization network for fine-tuning the learned classification network to localize each action instance. Escorcia et al. \cite{escorcia2016daps} introduce Deep Action Proposals (DAPs) and use a LSTM to encode information in a fixed clip (512 frames) of a video. After encoding information in the video clip, the LSTM scores K (64) predefined start and end positions in that clip. The start and end positions are selected based on statistics of the video dataset. We show that our method performs better than global representations like LSTMs which create a single feature representation for all scales in a video for localization of activities. In contemporary work, Shou et al. \cite{cdc_shou_cvpr17}  proposed a convolutional-de-convolutional (CDC) network by combing temporal upsampling and spatial downsampling for activity detection. Such an architecture helps in precise localization of activity boundaries. We show that the activity proposals generated by our method can further improve CDC's performance.

Context has been widely used in various computer vision algorithms. For example, it helps in tasks like object detection \cite{gidaris2015object}, semantic segmentation \cite{mottaghi2014role}, referring expressions \cite{yu2016modeling} etc. In videos it has been used for action and activity recognition \cite{hasan2015context,wu2011action}. However, for temporal localization of activities, existing methods do not employ temporal context, which we show is critical for solving this problem.

%Although it is possible to detect the presence of an activity by observing a short part of a video, predicting the beginning and end requires us to observe both these events. Current algorithms which generate temporal segments in videos use features inside the segment to construct its feature representation, and foreground/background labels are assigned based on its overlap with ground truth segments \cite{escorcia2016daps, caba2016fast,yuan2016temporal}. 

%\begin{figure}[tb]
%    \center
%    \includegraphics[width=0.7\linewidth]{figures%/figure_0.png}
%    \caption{ A visual depiction of our proposal generation algorithm. The green boxes represent K different proposals which are placed at a fixed interval in a video.}
%    \label{fig:pyramid}
%\end{figure}

\section{Approach}
Given a video $\mathcal{V}$, consisting of $T$ frames, TCN generates a ranked list of segments $s_1, s_2, ..., s_N$, each associated with a score. Each segment $s_j$ is a tuple $t_b, t_e$, where $t_b$ and $t_e$ denote the beginning and end of a segment. For each frame, we compute a $D$ dimensional feature vector representation which is generated using a deep neural network. An overview of our method is shown in Figure \ref{fig:arch}.

\subsection{Proposal Generation}
Our goal in this step is to use a small number of proposals to obtain high recall. First, we employ a temporal sliding window of a fixed length of $L$ frames with 50\% overlap. Suppose each video $\mathcal{V}$ has $M$ window positions. For each window at position $i$ ($i \in [0, M]$), its duration is specified as a tuple $(b_i, e_i)$, where $b_i$ and $e_i$ denote the beginning and end of a segment. We then, generate $K$ proposal segments (at $K$ different scales) at each position $i$. For $k \in [1, K]$, the segments are denoted by $(b_i^k, e_i^k)$. Also, the duration of each segment, $L_k$,  increases as a power of two, i.e $L_{k+1} = 2L_k$. This allows us to cover all candidate activity locations that are likely to contain activities of interests, and we refer them as activity proposals, $P=\{(b_i^k, e_i^k)\}_{i=0, k=1}^{M, K}$. Figure \ref{fig:demo} illustrates temporal proposal generation. When a proposal segment meets the boundary of a video, we use zero-padding. 

\subsection{Context Feature Representation}
We next construct a feature representation for ranking proposals. We use all the features $\mathcal{F} = \{f_1, f_2, ..., f_m\}$ of the untrimmed video as a feature representation for the video. For the $k^{th}$ proposal at window position $i$ ($P_{i,k}$), we uniformly sample from $\mathcal{F}$ to obtain a $D$ dimensional feature representation $Z_{i,k} = \{z_1, z_2, ..., z_n\}$. Here, $n$ is the number of features which are sampled from each segment. To capture temporal context, we again uniformly sample features from $\mathcal{F}$, but this time, from $P_{i,k+1}$ --- the proposal at the next scale and centered at the same scale. Note that we do not perform average or max-pooling but instead sample a fixed number of frames regardless of the duration of $P_{i,k}$.

Logically, a proposal can fall into one of four categories:

\begin{itemize}
    \item It is disjoint from a ground-truth interval and therefore, the next scale's (larger) label is irrelevant  
    \item It includes a ground-truth interval and the next-scale has partial overlap with that ground truth interval.
    \item It is included in a ground-truth interval and the next level has significant overlap with the background (i.e., it is larger than the ground truth interval).
    \item It is included in a ground-truth interval and so is the next level.   
\end{itemize}

A representation which only considers features inside a proposal would not consider the last two cases. Hence, whenever a proposal is inside an activity interval, it would not be possible to determine where the activity ends by only considering the features inside the proposal. Therefore, using a context based representation is critical for temporal localization of activities. Additionally, based on how much background the current and next scales cover, it becomes possible to determine if a proposal is a good candidate.

% \subsection{Pooling}
% After generating the feature pyramids of each video, we exploit pooling to concatenate all the features as $\mathcal{F}$. Here we use four different pooling methods as shown in Fig. \ref{fig:pyramid}.

\begin{figure*}
    \center
    \includegraphics[width=0.7\linewidth]{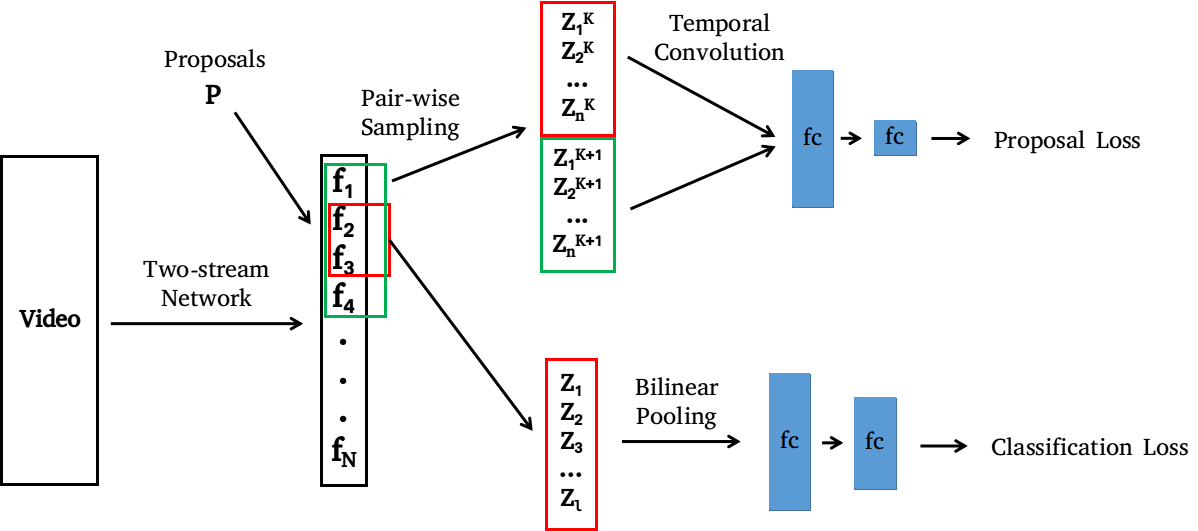}
    \caption{Temporal Context Network applies a two stream CNN on a video for obtaining an intermediate feature representation.}
    \label{fig:arch}
\end{figure*}

\subsection{Sampling and Temporal Convolution}
To train the proposal network, we assign labels to proposals based on their overlap with ground truth, as follows,
\begin{equation}
Label(S_j) = 
  \begin{cases}
  1, &    iou(S_j, GT) > 0.7  \\
  0, &    iou(S_j, GT) < 0.3  \\
  \end{cases}
\end{equation}
where $iou(\cdot)$ is intersection over union overlap and $GT$ is a ground truth interval. During training, we construct a mini batch with 1024 proposals with a positive to negative ratio of 1:1. 

Given a pair of features $Z_{i,k}$, $Z_{i,k+1}$, from two consecutive scales, we apply temporal convolution to features sampled from each temporal scale separately to capture context information between scales, as shown in Figure \ref{fig:arch}. A temporal Convolutional Neural Network \cite{kang2016object} enforces temporal consistency and obtains consistent performance improvements over still-image detections. To aggregate information across scales, we concatenate the two features to obtain a fixed dimensional representation. Finally, two fully connected layers are used to capture context information across scales. A two-way Softmax layer followed by cross-entropy loss is used at the end to map the predictions to labels (proposal or not).

\subsection{Classification}
Given a proposal with a high score, we need to predict its action class. We use bilinear pooling by computing the outer product of each segment feature, and average pool them to obtain the bilinear matrix $bilinear(\cdot)$. Given features $\hat{Z} = [z_1, z_2, ... z_l]$ within a proposal, we conduct bilinear pooling as follows:

	    \begin{equation}
			bilinear(\hat{Z}) = \sum_{i=1}^{l}\hat{Z}_{i}^T \hat{Z}_{i}
	    \end{equation}  

For classification, we pool all the features $l$ which are inside the segment and do not perform any temporal sampling. We pass this vectorized bilinear feature $x = bilinear(\hat{Z}) $ through a mapping function with signed square root and $l^2$ normalization \cite{ifv}:
	    \begin{equation}
	    \phi(x) = \frac{sign(x)\sqrt{x}}{||sign(x)\sqrt{x}||_2}
	    \end{equation}   

We finally apply a fully connected layer and use a 201-way (200 action classes plus background) Softmax layer at the end to predict class labels. We again use the cross entropy loss function for training. During training, we sample 1024 proposals to construct a mini batch. To balance training, 64 samples are selected as background in each mini-batch. For assigning labels to video segments, we use the same function which is used for generating proposals, 

\begin{equation}
Label(S_j) = 
  \begin{cases}
  lb, &    iou(S_j, GT) > 0.7  \\
  0, &    iou(S_j, GT) < 0.3  \\
  \end{cases}
\end{equation}
where $iou(\cdot)$ is intersection over union overlap, $GT$ is ground truth and $lb$ is the most dominant class with in proposal $S_j$. We use this classifier for the ActivityNet dataset but this can be replaced with other classifiers as well.

\begin{figure*}[t]
    \center
    \includegraphics[width=0.9\linewidth]{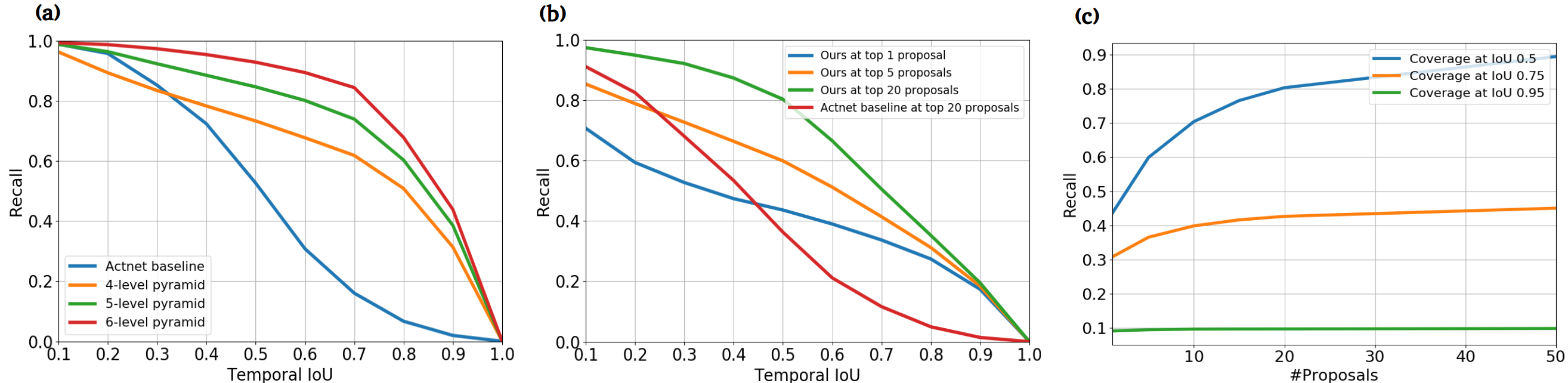}
    \caption{Performance of our proposal ranker on ActivityNet validation set. (a) The Recall vs IoU for pyramid proposal anchors; (b) The Recall vs IoU for our ranker at 1, 5, 20 proposals; (c) Recall vs number of proposals for our ranker at IoU 0.5, 0.75 and 0.95}
    \label{fig:ranker}
\end{figure*}

\begin{figure*}[t]
    \center
    \includegraphics[width=0.9\linewidth]{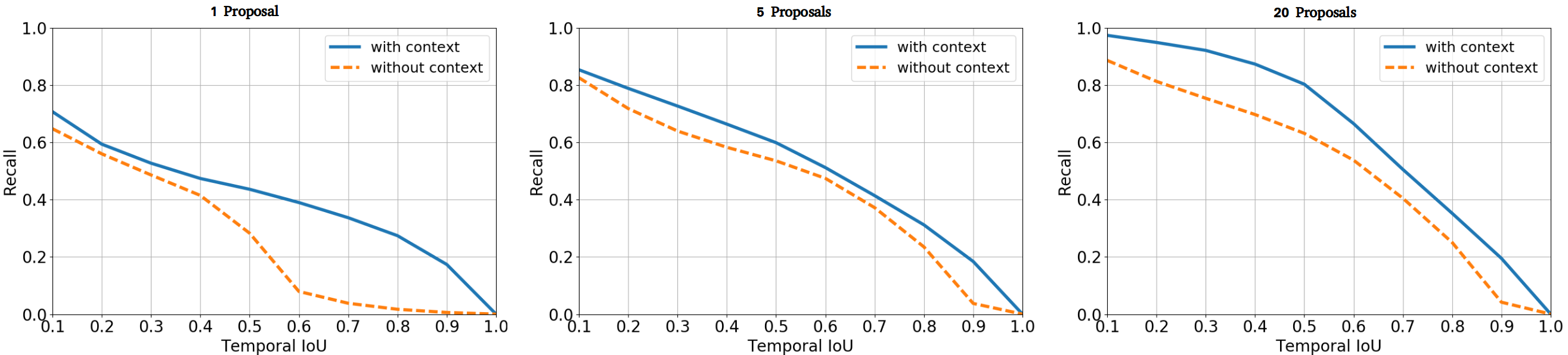}
    \caption{The effectiveness of context-based proposal ranker is shown in these plots. The Recall vs IoU plots show ranker performance at 1, 5, 20 proposals with and without context on ActivityNet validation set}
    \label{fig:withpair}
\end{figure*}

\begin{figure*}[t]
    \center
    \includegraphics[width=0.9\linewidth]{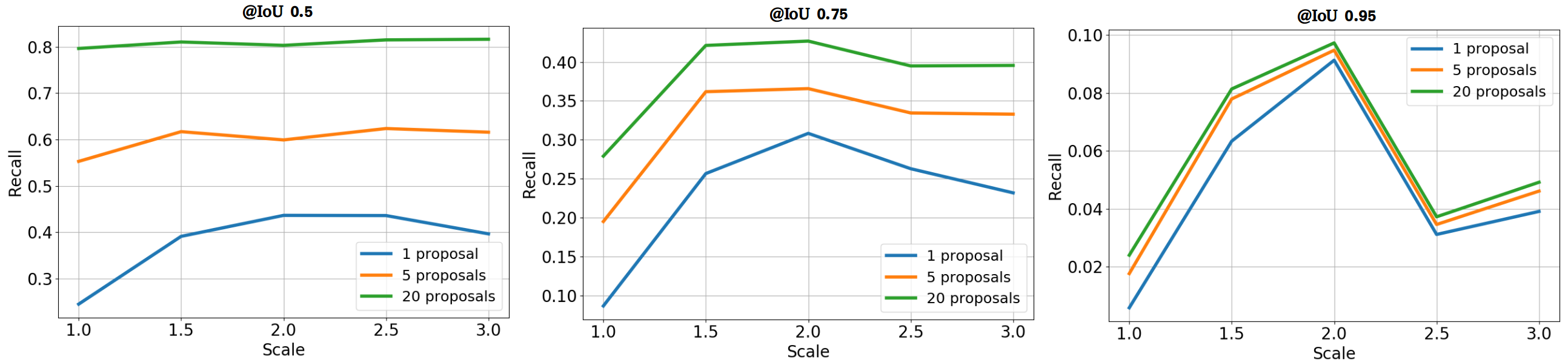}
    \caption{Comparing the ranker performance using different relative scale for context based proposals on ActivityNet validation set}
    \label{fig:scale}
\end{figure*}

\section{Experiments}
In this section, we provide analysis of our proposed temporal context network. We perform experiments on the ActivityNet and THUMOS14 datasets.

\subsection{Implementation details}
We implement the network based on a customized Caffe repository with Python interface. All evaluation experiments are performed on a workstation with a Titan X (Maxwell) GPU. We initialize our network with pre-trained TSN models \cite{wang2016temporal} and fine-tune them on both action labels and foreground/background labels to capture ``actionness" and "backgroundness". Later, we concatenate these together as high-level features input to our proposal ranker and classifier. For the proposal ranker, we use temporal convolution with a kernel size of 5 and a stride of 1, followed by ReLU activation and average pooling with size 3 and stride 1. The temporal convolution responses are then concatenated and mapped to a fully connected layer with 500 hidden units, which is used to predict the proposal score. To evaluate our method on the detection task, we generate top K proposals (K is set to 20, we apply non-maximum suppression to filter out similar proposals, using an NMS threshold set as 0.45) and classify them separately. While classifying proposals, we also fuse two global video level priors using ImageNet shuffle features \cite{ImagenetShuffle} and ``actionness" features to further improve classification performance, as shown in \cite{SinghC16}. We also perform an ablation study for different components of classification. For training the proposal network, we use a learning rate 0.1. For the classification network, we set learning the rate to 0.001. For both cases, we use a momentum of 0.9 and 5e-5 weight decay.

\subsection{ActivityNet Dataset}
ActivityNet \cite{caba2015activitynet} is a recently released dataset which contains 203 distinct action classes and a total of 849 hours of videos collected from YouTube. It consists of both trimmed and untrimmed videos. Each trimmed video contains a specific action with annotated segments. Untrimmed videos contain one or more activities with background involved. On average, each activity category has 137 untrimmed videos. Each video on average has 1.41 activities which are annotated with beginning and end points. This benchmark is designed for three applications: untrimmed video classification, trimmed activity classification, and untrimmed activity detection. Here, we evaluate our performance on the detection task in untrimmed videos. We use the mean average precision (mAP) averaged over multiple overlap thresholds to evaluate detection performance. Since test labels of ActivityNet are not released, we perform ablation studies on the validation data and test our full model on the evaluation server.

% \subsubsection{Proposal Ranker Analysis}

\textbf{Proposal anchors} We sample pair-wise proposals within a temporal pyramid. In Figure \ref{fig:ranker}(a), we present the recall for the pyramid proposal anchors on ActivityNet validation set with three different levels. This figure shows the theoretical best recall one can obtain using such a pyramid. Notice that even with a 4-level pyramid with 64 proposals in total, the coverage is already better than the baseline provided in the challenge, which uses 90 proposals. This ensures our proposal ranker's performance is high with a low number of proposals.

\textbf{Performance of our ranker} We evaluate our ranker with different numbers of proposals. Figure \ref{fig:ranker}(b) shows the average recall at various overlap thresholds with top 1, top 5 and top 20 proposals. Even when using {\em one} proposal, our ranker outperforms the ActivityNet proposal baseline by a significant margin when the overlap threshold is greater than 0.5. With top 20 proposals, our ranker can squeeze out most of the performance from pyramid proposal anchors. We also evaluate the performance of our ranker by measuring recall as the number of proposals varies (shown in Figure \ref{fig:ranker}(c)). Recall at IoU 0.5 increases to 90\% with just 20 proposals. At higher IoU, increasing the number of proposals does not increase recall significantly.

\begin{table}[]
    \centering
    \begin{tabular}{c|c|c|c}
        \hline
         & mAP@.5 & mAP@.75 & mAP@.95 \\
        \hline
        without context & 15.91 & 3.11 & 0.13 \\
        with context & 36.17 & 21.12 & 3.89 \\
        \hline
    
    \end{tabular}
    \caption{Evaluation on the influence with and without context on ActivityNet validation set}
    \label{tab:withpair}
\end{table}

\textbf{Effectiveness of temporal context}
We contend that temporal context for ranking proposals is critical for localization. To evaluate this claim, we conduct several experiments. In Figure \ref{fig:withpair}, we compare the performance of the ranker with and without temporal context. Using only the best proposal, without context, the recall drops significantly at high IoU (IoU $>$ 0.5). This shows that for precise localization of boundaries, temporal context is critical. Using top 5 and top 20 proposals, without context, the recall is marginally worse. This is expected because as the number of proposals increases, there is a higher likelihood of one having a good overlap with a ground-truth. Therefore, recall results using a single proposal are most informative. We also compute detection metrics on the ActivityNet validation set to evaluate the influence of context. Table \ref{tab:withpair} also shows that detection mAP is much higher when using the ranker with context based proposals. These experiments demonstrate the effectiveness of our method.

\textbf{Varying context window for ranking proposals}
Another important component for ranking proposals is the scale of context features which are associated with the proposal. Consider a case in which a proposal is contained within the ground truth interval. If the context scale is large, the ranker may not be able to distinguish between good and bad proposals, as it always see a significant amount of background . If the scale is small, there may be not enough context to determine if the proposal is contained within the ground truth or not. Therefore, we conduct an experimental study by varying the scale of context features while ranking proposals. In Figure \ref{fig:scale}, we observe that the performance improves up to a scale of 2. We evaluate the performance of the ranker at different scales on the ActivityNet validation set. In Table \ref{tab:scale} we show the impact of varying temporal context at different overlap thresholds, which validates our claim that adding more temporal context would hurt performance, but not using context at all would reduce performance by a much larger margin. For example, changing the scale from 2 to 3 only drops the performance by 3\% but changing it from 1.5 to 1 decreases mAP by 15\% and 12\% respectively. 

\begin{table}[]
    \centering
    \begin{tabular}{c|c|c|c}
        \hline
        Context Scale & mAP@.5 & mAP@.75 & mAP@.95 \\
        \hline
        1 & 15.91 & 3.11 & 0.13 \\
        1.5 & 30.51 & 15.56 & 2.23 \\
        2 & 36.17 & 21.12 & 3.89 \\
        2.5 & 36.04 & 17.08 & 0.92 \\
        3 & 33.29 & 14.35 & 1.03 \\
        \hline
    
    \end{tabular}
    \caption{Impact of varying temporal context at different overlap thresholds on ActivityNet validation set}
    \label{tab:scale}
\end{table}

% \subsubsection{Proposal Classifier Analysis}

\begin{table}[]
    \centering
    \begin{tabular}{c|c|c|c}
        \hline
        \#Proposal/Video & mAP@.5 & mAP@.75 & mAP@.95 \\
        \hline
        1 & 25.70 & 16.08 & 2.80 \\
        5 & 34.13 & 20.72 & 3.89 \\
        10 & 35.52 & 21.02 & 3.89 \\
        20 & 36.17 & 21.12 & 3.89 \\
        50 & 36.44 & 21.15 & 3.90 \\
        \hline
    \end{tabular}
    \caption{Impact of number proposals on mAP on ActivityNet validation set}
    \label{tab:topk}
\end{table}

\begin{table}[]
    \centering
    \begin{tabular}{c|c|c|c|c|c}
        \hline
        \multicolumn{3}{c|}{Components}  & mAP@.5 & mAP@.75 & mAP@.95 \\
        \multicolumn{1}{c}{B.} & \multicolumn{1}{c}{F.} & \multicolumn{1}{c}{G.} & \multicolumn{1}{|c}{} & \multicolumn{1}{|c}{} & \multicolumn{1}{|c}{} \\
        \hline
        \checkmark & \checkmark & \checkmark & 36.17 & 21.12  & 3.89 \\
        \checkmark & \checkmark & $\times$ & 33.83 & 20.05 & 3.77\\
        \checkmark & $\times$ & $\times$ & 30.31 & 17.80 & 2.82\\
        $\times$ & $\times$ & $\times$ & 26.35  & 15.27 & 2.66\\
        \hline
    
    \end{tabular}
    \caption{Ablation study for detection performance using top 20 proposals on the ActivityNet validation set. B - Bilinear, F - Flow, G - Global prior}
    \label{tab:abl}
\end{table}

\textbf{Influence of number of proposals}
We also evaluate the influence of the number of proposals on detection performance. Table \ref{tab:topk}, shows that our method doesn't requires a large number of proposals to improve its highest mAP. This demonstrates the advantages of both our proposal ranker and classifier.

\textbf{Ablation study}
We conduct a series of ablation studies to evaluate the importance of each component used in our classification model. Table \ref{tab:abl} considers three components: "B" stands for ``using bilinear pooling"; ``F" stands for ``using flow" and ``G" stands for ``using global priors". We can see from the table that each component plays a significant role in improving performance.

 \begin{table}[]
     \centering
     \small
     \begin{tabular}{c|c|c|c|c}
         \hline
         \multicolumn{5}{c}{Evaluation Server} \\
         \hline
         Method & mAP@.5 & mAP@.75 & mAP@.95 & Average \\
         \hline
         QCIS\cite{WangUTS} & 42.48 & 2.88 & 0.06 & 14.62\\
         UPC\cite{Montes_2016_NIPSWS} & 22.37 & 14.88 & 4.45 & 14.81\\
         UMD\cite{Singh_2016_CVPR} & 28.67 & 17.78 & 2.88 & 17.68\\
         Oxford\cite{SinghC16} & 36.40 & 11.05 & 0.14 & 17.83\\
         \hline
         \textbf{Ours} & \textbf{37.49} & \textbf{23.47} & \textbf{4.47} & \textbf{23.58} \\  
         \hline
    
     \end{tabular}
     \caption{Comparison with state-of-the-art methods on the ActivityNet evaluation sever using top 20 proposals}
     \label{tab:stoa}
 \end{table}

\textbf{Comparison with state-of-the-art}
 We compare our method with state-of-the-art methods \cite{WangUTS, Montes_2016_NIPSWS, SinghC16, SinghC16} submitted during the CVPR 2016 challenge. We submit our results on the evaluation server to measure performance on the test set. At 0.5 overlap, our method is only worse than \cite{WangUTS}. However, this approach was optimized for 0.5 overlap and its performance degrades significantly (to 2\%) when mAP at 0.75 or 0.95 overlap is measured. Even though frame level predictions using a Bi-directional LSTM are used in \cite{Singh_2016_CVPR}, our performance is better when mAP is measured at 0.75 overlap. This is because \cite{Singh_2016_CVPR} only performs simple grouping of contiguous segments which are obtained at multiple detection thresholds, instead of a proposal based approach. Hence, it is likely to perform worse on longer action segments.

\subsection{The THUMOS14 Dataset}

We also evaluate our framework on the THUMOS14 dataset\cite{jiang2014thumos}, which contains 20 action categories from sports. The validation set contains 1010 untrimmed videos with 200 videos as containing positive samples. The testing set contains 1574 untrimmed videos, where only 213 of them have action instances. We exclude the remaining background videos from our experiments.

Note that solutions for action and activity detection could be different in general, as activities could be very long (minutes) while actions last just a few seconds. Due to their long duration, evaluation at high overlap (0.8 e.g.) makes sense for activities, but not for actions. Nevertheless, we also train our proposed framework on the validation set of THUMOS14 and test on the testing set. Our model also outperforms state-of-the-art methods on proposal metrics by a significant margin, which shows the good generalization ability of our approach.

\textbf{Performance of our ranker}
Our proposal ranker outperforms existing algorithms like SCNN\cite{scnn_shou_wang_chang_cvpr16} and DAPs\cite{escorcia2016daps}. We show proposal performance on both average recall calculated using IoU thresholds from 0.5 to 1 at a step 0.05 (shown in Table \ref{tab:thumos_proposal1}) and recall at IoU 0.5 (shown in Table \ref{tab:thumos_proposal2}) using 10, 50, 100, 500 proposals. Our proposal ranker performs consistently better than previous methods, especially using small number of proposals.

In Table \ref{tab:thumos_proposal3}, it is clear that, the proposal ranker performance improves significantly when using a pair of context windows as input. Hence, it is important to use context features for localization in videos, which has been largely ignored in previous state-of-the-art activity detection methods.

\textbf{Comparison with state-of-the-art}
Using off the shelf classifiers and our proposals, we also demonstrate noticeable improvement in detection performance on THUMOS14. Here, we compare our temporal context network with DAPs\cite{escorcia2016daps}, PSDF\cite{psdf_cvpr16}, FG\cite{fg_cvpr16} SCNN\cite{scnn_shou_wang_chang_cvpr16} and CDC\cite{cdc_shou_cvpr17}. We replace the S-CNN proposals originally used in CDC with our proposals. For scoring the detections in CDC, we multiply our proposal scores with CDC's classification score. We show that our proposals further benefit CDC and improve detection performance consistently at different overlap thresholds.

\begin{table}[]
    \centering
    \begin{tabular}{c|c|c|c|c}
    \hline
    \multirow{2}{*}{Method}& \multicolumn{4}{c}{Avg.Recall [0.5:0.05:1]} \\
    & \multicolumn{1}{c}{@10}& \multicolumn{1}{c}{@50}&	\multicolumn{1}{c}{@100}&	\multicolumn{1}{c}{@500}\\ 
    \hline
    DAPs&	3.0&	11.7&	20.1&	46.7\\
    SCNN&	5.5&	16.6&	24.8&	48.3\\
    Ours&	7.7&	20.5&	29.6&	49.2\\ 
    \hline
    \end{tabular}
    \caption{Average Recall from IoU 0.5 to 1 with step size 0.05 for our proposals and other methods on the THUMOS14 testing set}
    \label{tab:thumos_proposal1}
\end{table}

\begin{table}[]
    \centering
    \begin{tabular}{c|c|c|c|c}
    \hline    
    \multirow{2}{*}{Method}& \multicolumn{4}{c}{Recall(IoU=0.5)}\\
    &\multicolumn{1}{c}{@10}& \multicolumn{1}{c}{@50}& \multicolumn{1}{c}{@100}&	\multicolumn{1}{c}{@500}\\ 
    \hline
    DAPs&	8.4&    29.2&   46.9&	85.5\\
    SCNN&	13.0&	35.2&	49.6&	84.1\\
    Ours&	17.1&	42.8&	59.8&	88.7\\
    \hline
    \end{tabular}
    \caption{Recall evaluation at IoU 0.5 between our proposals and state-of-the-art methods on THUMOS14 testing set}
    \label{tab:thumos_proposal2}
\end{table}

\begin{table}[]
    \centering
    \begin{tabular}{c|c|c}
    \hline 
    Method&	    Avg.Recall@100&	mAP@0.5\\
    \hline
    Ours w/o Context&	22.5&	20.5\\
    Ours w/ Context&	29.6&	25.6\\
    \hline
    \end{tabular}
    \caption{Evaluation on the influence with and without context on THUMOS14 testing set}
    \label{tab:thumos_proposal3}
\end{table}

\begin{table}[]
    \centering
    \small
    \begin{tabular}{c|c|c|c|c}
    \hline  
    Method&	mAP@.4&	mAP@.5&	mAP@.6&	mAP@.7\\
    \hline
    DAPs\cite{escorcia2016daps}&	----&	13.9&	----&	----\\ 
    FG\cite{fg_cvpr16}&     26.4&   17.1&   ----&   ----\\
    PSDF\cite{psdf_cvpr16}&	26.1&	18.8&	----&	----\\
    SCNN\cite{scnn_shou_wang_chang_cvpr16}&	28.7&	19.0&	----&	----\\
SCNN+CDC\cite{cdc_shou_cvpr17}&	29.4&	23.3&	13.1&	7.9\\
    \hline
    \textbf{Ours}+CDC&	\textbf{33.3}&	\textbf{25.6}&	\textbf{15.9}&	\textbf{9.0}\\
    \hline
    \end{tabular}
    \caption{Performance of state-of-the-art detectors on the THUMOS14 testing set}
    \label{tab:thumos_detection}
\end{table}

\begin{figure*}[t]
    \center
    \includegraphics[width=0.85\linewidth]{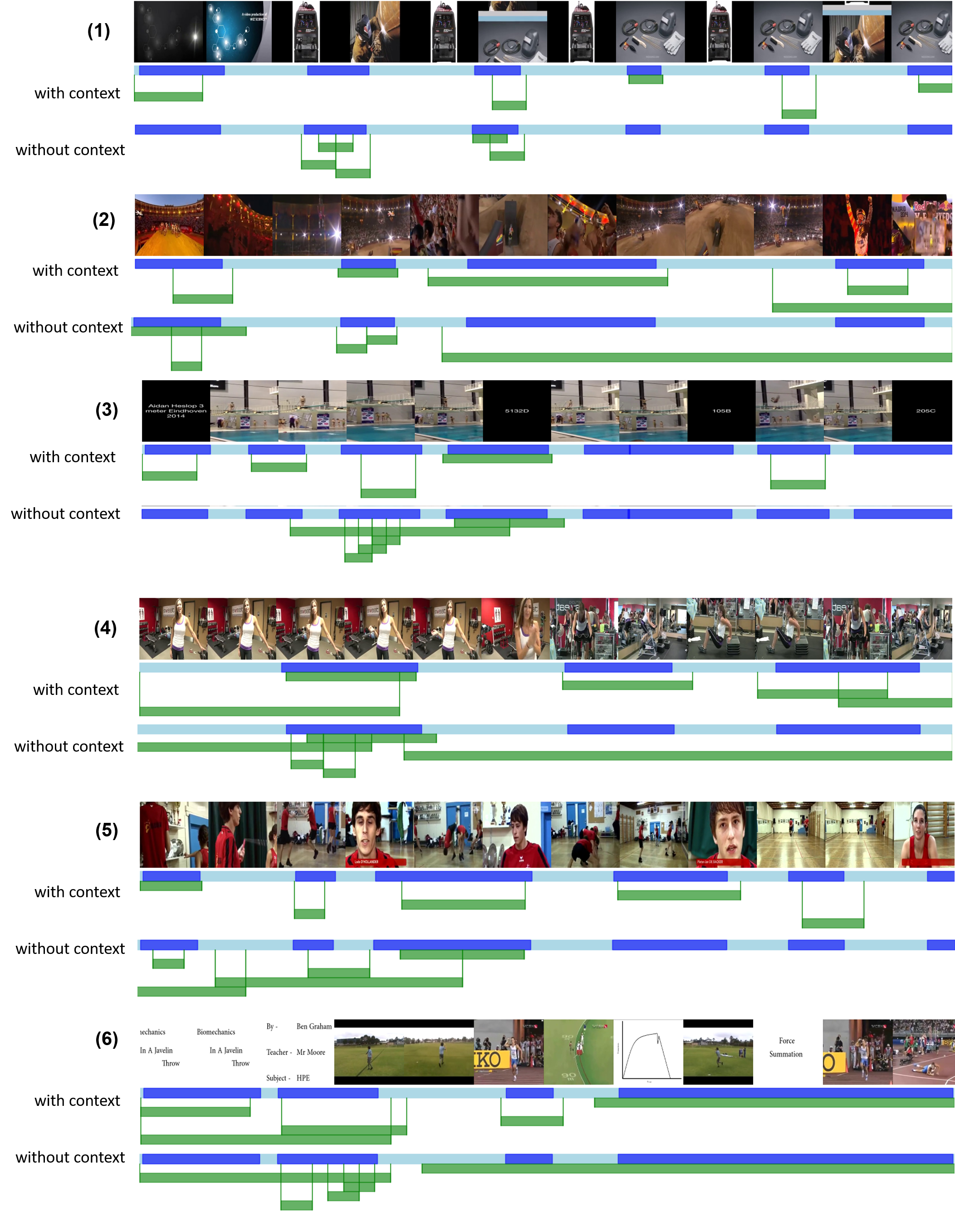}
    \caption{Visualization of top 5 ranking results, the blue bar denotes the ground-truth while the green one represents proposals.}
    \label{fig:viz}
\end{figure*}

\section{Qualitative Results}
We show some qualitative results for TCN, with and without context. Note that only top 5 proposals are shown. The ground truth is shown in blue while predictions are shown in green. It is evident that when context is not used, multiple proposals are present inside or just at the boundary of ground truth intervals. Therefore, although the location is near the actual interval (when context is not used), the boundaries are inaccurate. Hence, when detection metrics are computed, these nearby detections get marked as false positives leading to a drop in average precision.  However, when context is used, the proposals boundaries are significantly more accurate compared to the case when context is not used. 
%We also show two failure cases when context is used. These are due to very little movement (6) or an activity that extends beyond the video (5).

\section{Conclusion}
We demonstrated that temporal context is helpful for performing localization of activities in videos. Analysis was performed to study the impact of temporal proposals in videos by studying precision recall characteristics at multiple overlap thresholds. We also vary the context window to study the importance of temporal context for localization. Finally, we demonstrated state-of-the-art performance on two challenging public datasets.

\section*{Acknowledgement}
\vspace*{-0.1cm}
The authors acknowledge the University of Maryland supercomputing resources \url{http://www.it.umd.edu/hpcc} made available for conducting the research reported in this paper.

{\small
\bibliographystyle{ieee}
\bibliography{egbib}

\begin{thebibliography}{10}\itemsep=-1pt

\bibitem{atmosukarto2012trajectory}
I.~Atmosukarto, B.~Ghanem, and N.~Ahuja.
\newblock Trajectory-based fisher kernel representation for action recognition
  in videos.
\newblock In {\em Pattern Recognition (ICPR), 21st International Conference
  on}, pages 3333--3336. IEEE, 2012.

\bibitem{bobick2001recognition}
A.~F. Bobick and J.~W. Davis.
\newblock The recognition of human movement using temporal templates.
\newblock {\em IEEE Transactions on pattern analysis and machine intelligence},
  23(3):257--267, 2001.

\bibitem{bregler1997learning}
C.~Bregler.
\newblock Learning and recognizing human dynamics in video sequences.
\newblock In {\em Computer Vision and Pattern Recognition (CVPR), IEEE
  Conference on}, pages 568--574. IEEE, 1997.

\bibitem{caba2016fast}
F.~Caba~Heilbron, J.~Carlos~Niebles, and B.~Ghanem.
\newblock Fast temporal activity proposals for efficient detection of human
  actions in untrimmed videos.
\newblock In {\em Computer Vision and Pattern Recognition (CVPR), IEEE
  Conference on}, pages 1914--1923, 2016.

\bibitem{caba2015activitynet}
F.~Caba~Heilbron, V.~Escorcia, B.~Ghanem, and J.~Carlos~Niebles.
\newblock Activitynet: A large-scale video benchmark for human activity
  understanding.
\newblock In {\em Proceedings of the IEEE Conference on Computer Vision and
  Pattern Recognition}, pages 961--970, 2015.

\bibitem{escorcia2016daps}
V.~Escorcia, F.~C. Heilbron, J.~C. Niebles, and B.~Ghanem.
\newblock Daps: Deep action proposals for action understanding.
\newblock In {\em European Conference on Computer Vision}, pages 768--784.
  Springer, 2016.

\bibitem{gemert2015apt}
J.~Gemert, M.~Jain, E.~Gati, C.~G. Snoek, et~al.
\newblock {\em Apt: Action localization proposals from dense trajectories}.
\newblock BMVA Press, 2015.

\bibitem{gidaris2015object}
S.~Gidaris and N.~Komodakis.
\newblock Object detection via a multi-region and semantic segmentation-aware
  cnn model.
\newblock In {\em Proceedings of the IEEE International Conference on Computer
  Vision}, pages 1134--1142, 2015.

\bibitem{gkioxari2015finding}
G.~Gkioxari and J.~Malik.
\newblock Finding action tubes.
\newblock In {\em Computer Vision and Pattern Recognition (CVPR), IEEE
  Conference on}, pages 759--768, 2015.

\bibitem{haritaoglu2000w}
I.~Haritaoglu, D.~Harwood, and L.~S. Davis.
\newblock W/sup 4: real-time surveillance of people and their activities.
\newblock {\em IEEE Transactions on pattern analysis and machine intelligence},
  22(8):809--830, 2000.

\bibitem{hasan2015context}
M.~Hasan and A.~K. Roy-Chowdhury.
\newblock Context aware active learning of activity recognition models.
\newblock In {\em Proceedings of the IEEE International Conference on Computer
  Vision}, pages 4543--4551, 2015.

\bibitem{heilbron2014camera}
F.~C. Heilbron, A.~Thabet, J.~C. Niebles, and B.~Ghanem.
\newblock Camera motion and surrounding scene appearance as context for action
  recognition.
\newblock In {\em Asian Conference on Computer Vision}, pages 583--597.
  Springer, 2014.

\bibitem{jain2014action}
M.~Jain, J.~Van~Gemert, H.~J{\'e}gou, P.~Bouthemy, and C.~G. Snoek.
\newblock Action localization with tubelets from motion.
\newblock In {\em Computer Vision and Pattern Recognition (CVPR), IEEE
  Conference on}, pages 740--747, 2014.

\bibitem{jiang2014thumos}
Y.~Jiang, J.~Liu, A.~R. Zamir, G.~Toderici, I.~Laptev, M.~Shah, and
  R.~Sukthankar.
\newblock Thumos challenge: Action recognition with a large number of classes,
  2014.

\bibitem{psdf_cvpr16}
X.~Y. A.~A. Jun~Yuan, Bingbing~Ni.
\newblock Temporal action localization with pyramid of score distribution
  features.
\newblock In {\em Computer Vision and Pattern Recognition (CVPR), IEEE
  Conference on}, 2016.

\bibitem{kang2016object}
K.~Kang, W.~Ouyang, H.~Li, and X.~Wang.
\newblock Object detection from video tubelets with convolutional neural
  networks.
\newblock In {\em Computer Vision and Pattern Recognition (CVPR), IEEE
  Conference on}, pages 817--825, 2016.

\bibitem{ma2016learning}
S.~Ma, L.~Sigal, and S.~Sclaroff.
\newblock Learning activity progression in lstms for activity detection and
  early detection.
\newblock In {\em Computer Vision and Pattern Recognition (CVPR), IEEE
  Conference on}, pages 1942--1950, 2016.

\bibitem{mettes2015bag}
P.~Mettes, J.~C. van Gemert, S.~Cappallo, T.~Mensink, and C.~G. Snoek.
\newblock Bag-of-fragments: Selecting and encoding video fragments for event
  detection and recounting.
\newblock In {\em Proceedings of the 5th ACM on International Conference on
  Multimedia Retrieval}, pages 427--434. ACM, 2015.

\bibitem{ImagenetShuffle}
P.~S.~M. Mettes, D.~C. Koelma, and C.~G.~M. Snoek.
\newblock The imagenet shuffle: Reorganized pre-training for video event
  detection.
\newblock In {\em ACM International Conference on Multimedia Retrieval}, 2016.

\bibitem{Montes_2016_NIPSWS}
A.~Montes, A.~Salvador, S.~Pascual, and X.~Giro-i Nieto.
\newblock Temporal activity detection in untrimmed videos with recurrent neural
  networks.
\newblock In {\em 1st NIPS Workshop on Large Scale Computer Vision Systems},
  December 2016.

\bibitem{mottaghi2014role}
R.~Mottaghi, X.~Chen, X.~Liu, N.-G. Cho, S.-W. Lee, S.~Fidler, R.~Urtasun, and
  A.~Yuille.
\newblock The role of context for object detection and semantic segmentation in
  the wild.
\newblock In {\em Computer Vision and Pattern Recognition (CVPR), IEEE
  Conference on}, pages 891--898, 2014.

\bibitem{oneata2014spatio}
D.~Oneata, J.~Revaud, J.~Verbeek, and C.~Schmid.
\newblock Spatio-temporal object detection proposals.
\newblock In {\em European conference on computer vision}, pages 737--752.
  Springer, 2014.

\bibitem{peng2016bag}
X.~Peng, L.~Wang, X.~Wang, and Y.~Qiao.
\newblock Bag of visual words and fusion methods for action recognition:
  Comprehensive study and good practice.
\newblock {\em Computer Vision and Image Understanding}, 150:109--125, 2016.

\bibitem{ifv}
F.~Perronnin, J.~S\'{a}nchez, and T.~Mensink.
\newblock Improving the fisher kernel for large-scale image classification.
\newblock In {\em Proceedings of the 11th European Conference on Computer
  Vision: Part IV}, ECCV'10, pages 143--156, Berlin, Heidelberg, 2010.
  Springer-Verlag.

\bibitem{ren2015faster}
S.~Ren, K.~He, R.~Girshick, and J.~Sun.
\newblock Faster r-cnn: Towards real-time object detection with region proposal
  networks.
\newblock In {\em Advances in neural information processing systems}, pages
  91--99, 2015.

\bibitem{russakovsky2015imagenet}
O.~Russakovsky, J.~Deng, H.~Su, J.~Krause, S.~Satheesh, S.~Ma, Z.~Huang,
  A.~Karpathy, A.~Khosla, M.~Bernstein, et~al.
\newblock Imagenet large scale visual recognition challenge.
\newblock {\em International Journal of Computer Vision}, 115(3):211--252,
  2015.

\bibitem{fg_cvpr16}
G.~M. L. F.-F. Serena~Yeung, Olga~Russakovsky.
\newblock End-to-end learning of action detection from frame glimpses in
  videos.
\newblock In {\em Computer Vision and Pattern Recognition (CVPR), IEEE
  Conference on}, 2016.

\bibitem{cdc_shou_cvpr17}
Z.~Shou, J.~Chan, A.~Zareian, K.~Miyazawa, and S.-F. Chang.
\newblock Cdc: Convolutional-de-convolutional networks for precise temporal
  action localization in untrimmed videos.
\newblock In {\em Computer Vision and Pattern Recognition (CVPR), IEEE
  Conference on}, 2017.

\bibitem{shou2016action}
Z.~Shou, D.~Wang, and S.~Chang.
\newblock Action temporal localization in untrimmed videos via multi-stage
  cnns.
\newblock In {\em Computer Vision and Pattern Recognition (CVPR), IEEE
  Conference on}, 2016.

\bibitem{scnn_shou_wang_chang_cvpr16}
Z.~Shou, D.~Wang, and S.-F. Chang.
\newblock Temporal action localization in untrimmed videos via multi-stage
  cnns.
\newblock In {\em Computer Vision and Pattern Recognition (CVPR), IEEE
  Conference on}, 2016.

\bibitem{Singh_2016_CVPR}
B.~Singh, T.~K. Marks, M.~Jones, O.~Tuzel, and M.~Shao.
\newblock A multi-stream bi-directional recurrent neural network for
  fine-grained action detection.
\newblock In {\em The IEEE Conference on Computer Vision and Pattern
  Recognition (CVPR)}, June 2016.

\bibitem{SinghC16}
G.~Singh and F.~Cuzzolin.
\newblock Untrimmed video classification for activity detection: submission to
  activitynet challenge.
\newblock {\em CoRR}, abs/1607.01979, 2016.

\bibitem{wang2011action}
H.~Wang, A.~Kl{\"a}ser, C.~Schmid, and C.-L. Liu.
\newblock Action recognition by dense trajectories.
\newblock In {\em Computer Vision and Pattern Recognition (CVPR), IEEE
  Conference on}, pages 3169--3176.

\bibitem{wang2013action}
H.~Wang and C.~Schmid.
\newblock Action recognition with improved trajectories.
\newblock In {\em Proceedings of the IEEE International Conference on Computer
  Vision}, pages 3551--3558, 2013.

\bibitem{wang2016temporal}
L.~Wang, Y.~Xiong, Z.~Wang, Y.~Qiao, D.~Lin, X.~Tang, and L.~Van~Gool.
\newblock Temporal segment networks: towards good practices for deep action
  recognition.
\newblock In {\em European Conference on Computer Vision}, pages 20--36.
  Springer, 2016.

\bibitem{WangUTS}
R.~Wang and D.~Tao.
\newblock Uts at activitynet 2016.
\newblock {\em AcitivityNet Large Scale Activity Recognition Challenge}, 2016.

\bibitem{wang2016improving}
Y.~Wang and M.~Hoai.
\newblock Improving human action recognition by non-action classification.
\newblock In {\em Computer Vision and Pattern Recognition (CVPR), IEEE
  Conference on}, pages 2698--2707, 2016.

\bibitem{wu2011action}
X.~Wu, D.~Xu, L.~Duan, and J.~Luo.
\newblock Action recognition using context and appearance distribution
  features.
\newblock In {\em Computer Vision and Pattern Recognition (CVPR), IEEE
  Conference on}, pages 489--496. IEEE, 2011.

\bibitem{yu2015fast}
G.~Yu and J.~Yuan.
\newblock Fast action proposals for human action detection and search.
\newblock In {\em Computer Vision and Pattern Recognition (CVPR), IEEE
  Conference on}, pages 1302--1311, 2015.

\bibitem{yu2016modeling}
L.~Yu, P.~Poirson, S.~Yang, A.~C. Berg, and T.~L. Berg.
\newblock Modeling context in referring expressions.
\newblock In {\em European Conference on Computer Vision}, pages 69--85.
  Springer, 2016.

\bibitem{yuan2016temporal}
J.~Yuan, B.~Ni, X.~Yang, and A.~A. Kassim.
\newblock Temporal action localization with pyramid of score distribution
  features.
\newblock In {\em Computer Vision and Pattern Recognition (CVPR), IEEE
  Conference on}, pages 3093--3102, 2016.

\end{thebibliography}
}

\end{document}